\pdfoutput=1

\documentclass[11pt]{article}

\usepackage[final]{acl}

\usepackage{times}
\usepackage{latexsym}

\usepackage[T1]{fontenc}

\usepackage[utf8]{inputenc}

\usepackage{microtype}

\usepackage{inconsolata}

\usepackage{booktabs}
\usepackage{graphicx}
\usepackage{subcaption}
\usepackage{newtxmath}

\title{\raisebox{-0.3\height}{\includegraphics[height=2.5em]{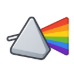}} \name{}: A Task-Agnostic Framework for Multi-Prompt Generation}

\usepackage{listings}
\usepackage{xcolor}
\usepackage{caption}
\usepackage{float}
\floatstyle{plain}
\newfloat{listing}{tb}{lop}
\floatname{listing}{Listing}

\definecolor{codegray}{gray}{0.95}

\lstdefinestyle{mystyle}{
    basicstyle=\ttfamily\small,
    keywordstyle=\color{blue},
    commentstyle=\color{gray},
    stringstyle=\color{red!70!black},
    numbers=none,
    breaklines=true,
    frame=tb,
    captionpos=b
}

\author{Eliya Habba\thanks{Equal contribution.} \quad Noam Dahan$^{*}$ \quad Gili Lior \quad Gabriel Stanovsky \\[1.5mm]
         The Hebrew University of Jerusalem \\[1.5mm] 
 \href{mailto:eliya.habba@mail.huji.ac.il}{\texttt{eliya.habba@mail.huji.ac.il}}}

\newcommand{\name}{PromptSuite}
\newcommand{\llama}{Llama-3.3-70B}
\newcommand{\gpt}{GPT-4o-mini}

\begin{document}
\maketitle
\begin{abstract}
Evaluating LLMs with a single prompt has proven unreliable, with small changes leading to significant performance differences. However, generating the prompt variations needed for a more robust multi-prompt evaluation is challenging, limiting its adoption in practice. To address this, we introduce \name{}, a framework that enables the automatic generation of various prompts. \name{} is flexible -- working out of the box on a wide range of tasks and benchmarks. It follows a modular prompt design, allowing controlled perturbations to each component, and is extensible, supporting the addition of new components and perturbation types.  Through a series of case studies, we show that \name{} provides meaningful variations to support strong evaluation practices. All resources, including the Python API, source code, user-friendly web interface, and demonstration video, are available at:
\noindent\url{https://eliyahabba.github.io/PromptSuite/}.

\end{abstract}

\section{Introduction}
\begin{figure*}[tb!]
    \centering
    \includegraphics[width=\textwidth]{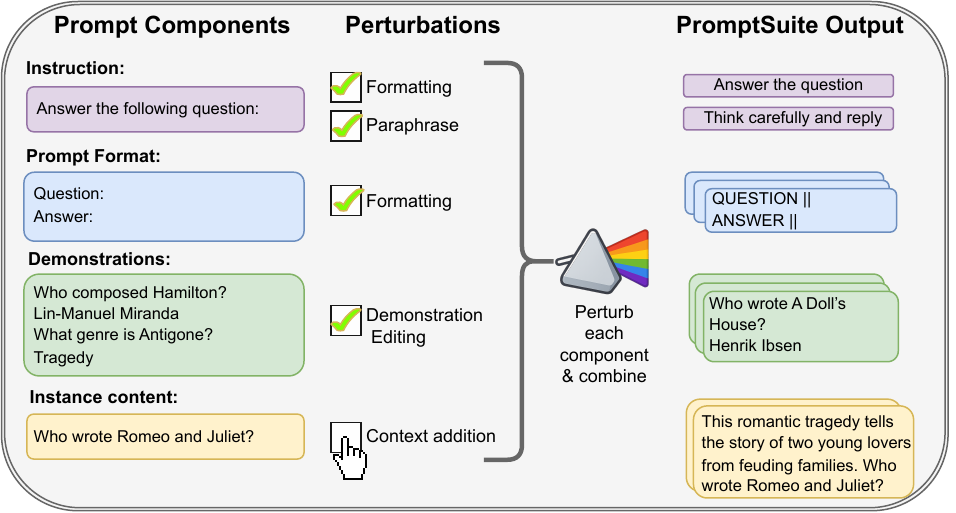}
\caption{\name{} framework: configure a modular prompt, and apply component-wise perturbations. This modularity enables \name{} to generalize across tasks and adapt to diverse data.}
\label{fig:just_componenet}
\end{figure*} 
Recent studies have demonstrated that LLMs are highly sensitive to small, meaning-preserving variations in task formulation. Minor changes, ranging from adding white spaces to instruction paraphrasing, lead to substantial differences in  performance and model ranking~\cite{sclar2023quantifying,mizrahi2024state}. 
This sensitivity has been explored in the evaluation of many NLP tasks in zero and few shot settings, such as text classifications~\cite{chakraborty2023zero,reif2024performancequantifyingmitigatinglabel}; multiple-choice question answering~\cite{habba2025dovelargescalemultidimensionalpredictions,alzahrani-etal-2024-benchmarks}; and text generation tasks~\cite{resendiz2024mopo}, raising concerns about the validity of evaluation performed using a single prompt. 

Evaluating over multiple prompts is currently challenging because there is no standard way to extend existing benchmarks, which were largely compiled using a single prompt. Evidently, despite its major limitation, single-prompt evaluations are still prevalent in many NLP tasks~\cite{gu-etal-2024-counterfeit, gu2024cruxeval,lior2025reliableeval}.





To address this major challenge standing in the way of meaningful evaluation in NLP, we present \name{}, a framework that generates multiple prompts, employing both LLMs as well as rule-based heuristics to generate variations along dimensions that were found to affect model performance. \name{} is built around three core principles, presented in Section~\ref{Sec:Principles}. First, \name{} is \emph{flexible}, designed to work out of the box on a wide range of benchmarks. Second, \name{} follows a \emph{modular} design  that decomposes prompts into four components: instruction, prompt format, demonstration, and instance content, and \name{} enables targeted perturbations to each component, making it easy to evaluate their impact and adapt to new tasks. Finally, \name{} is \emph{extensible,} supporting future LLM evaluation research with easy extensions for new prompt components and perturbations.

\name{} provides different types of perturbations to each prompt component, including formatting, paraphrasing, context addition, and few-shot demonstration editing, as illustrated in Figure~\ref{fig:just_componenet}. In Section~\ref{Sec:united}, we provide further details on the perturbation types, as well as demonstrate how to use our API to transform raw data into multiple prompt variations with just a few lines of code.

In Section~\ref{Sec:CaseStudies}, we demonstrate the flexibility of our framework through a series of case studies. We assess the impact of prompt variation on nine diverse tasks with two SOTA LLMs, highlighting the utility of \name{} for multi-prompt evaluation. 

Our contributions are as follows:
\begin{enumerate}
    \item We present \name{}, an easy-to-use framework that provides the prompt variations needed for multi-prompt evaluation across a wide range of tasks, working out of the box with diverse benchmarks and datasets. 
    \item We evaluate \name{}'s capabilities through a series of case studies that demonstrate its ability to reveal LLM sensitivity to prompt variations across a diverse set of tasks.

     \item We show \name{}'s modular design enables isolating the effect of individual prompt component perturbations, enabling future research into the causes of LLM sensitivity.

\end{enumerate}




    
    

\section{\name{}'s Principles}
\label{Sec:Principles}
We build \name{} on three core principles to make it useful across a variety of use cases and applicable over time.
\paragraph{Flexible} \name{} must be able to support a diverse range of tasks out of the box. We achieve this by relying only on the prompt structure and not its content. We demonstrate this flexibility in Section~\ref{Sec:CaseStudies}, where we use \name{} to get prompt variations in 9 different tasks, including question answering, multiple-choice reasoning, translation, summarization, and code generation.  

\paragraph{Modular} \name{} is built around a modular representation, treating each prompt as a combination of independent components, as shown in Figure~\ref{fig:just_componenet}. This design enables targeted perturbations to specific components, supporting evaluation of their impact on model performance and allowing  adaptation to new tasks, formats, or datasets.


\paragraph{Extensible} 
\name{} is built to support future research in the field of LLM evaluation. In particular, it is easy to add new prompt components and new types of perturbations through our open-source code.\footnote{\url{https://www.github.com/eliyahabba/PromptSuite
}} 






\section{\name{}}
\label{Sec:united}
\name{} is a flexible, modular and extensible framework that generates diverse prompts needed for robust evaluation. For a given dataset, it outputs a set of prompt variations for few-shot or zero-shot settings, where each sample from the dataset appears multiple times with different prompts. For example, in Listing~\ref{lst:api}, we load SQuAD~\cite{rajpurkar2016squad100000questionsmachine} through Hugging Face, set up the prompt and choose to paraphrase the instruction and apply formatting to the prompt format. This results in 9 prompt variations per sample, with just a few lines of code, and minimal information about the dataset. 

In this section, we briefly describe the modular prompt design and the supported perturbations, followed by an overview of how \name{} is used.

\begin{listing}[tb!]

\begin{lstlisting}[language=Python, showspaces=false, caption={Code snippet for using PromptSuite API. Here we apply paraphrasing to the instruction with an LLM and formatting to the prompt format. },label={lst:api}]
from promptsuite import PromptSuite

# Initialize
ps = PromptSuite()

# 1) Load dataset directly from HF
ps.load_dataset("rajpurkar/squad") 

# 2) Setup template and 3) Choose perturbations
template = {
  'instruction': 'Please answer the following questions.',
  'prompt format': 'Q: {question}\nA: {answer}',
"instruction variations": ["paraphrase_with_llm"],
"prompt format variations": ["format structure"],
}
ps.set_template(template)

# 4) Generate variations
ps.configure(variations_per_field=3,   api_platform="OpenAI", model_name="gpt-4o-mini")
variations = mp.generate(verbose=True)

# Export results
ps.export("output.json", format="json")
\end{lstlisting}
\end{listing}

\subsection{Modular Prompt and Perturbations}
\name{} treats each prompt as a concatenation of components, allowing controlled perturbations to each part, as illustrated in Figure~\ref{fig:just_componenet}. This interpretation of prompt structure is an integration of several recent works that identified prompt components that affect overall performance~\cite{sclar2023quantifying,mondshine2025beyond}.
Specifically, each prompt is comprised of: \emph{instruction} (e.g., ``Answer the following question'', ``Summarize the following text''); \emph{prompt format} (e.g, ``Question:, Answer:'', ``Text:, Summary:''); \emph{demonstrations}; and \emph{instance content} -- the current sample the model is evaluated on (``Who wrote Romeo and Juliet?'').  


\begin{table*}
\resizebox{\textwidth}{!}{%
\small
\centering
\begin{tabular}{@{}l p{3.5cm} p{7cm}@{}}

\toprule
\textbf{Perturbation Type} & \textbf{Applicable Components}  & \textbf{Description} \\
\midrule

Formatting & Instruction, prompt format, demonstrations, 

instance content & Adds surface-level noise to the text It includes inserting extra spaces, introducing typos, changing letter casing, and altering punctuation. Following~\cite{sclar2023quantifying}.
 \\ \hline

Paraphrase & Instruction & Creates semantically equivalent variations to the instruction that differ in phrasing and style. Following~\cite{mizrahi2024state}.  \\ \hline

Context addition & Instance content & Uses an LLM to add text related to the instance content without revealing or changing the answer. Following~\cite{liu2023lostmiddlelanguagemodels,levy2024same}. \\

\hline

Demonstration Editing & Demonstrations & Changes the few-shot examples, their order and their number. Following~\cite{lu2021fantastically}. \\

\bottomrule

\end{tabular}
}

\caption{Overview of the perturbation types supported by \name{}. The ``Applicable Components'' column specifies which prompt components each perturbation can be applied to. For example, paraphrasing is applicable to the instruction component. 
}
\label{tab:pert}
\end{table*}

Each component can be subjected to different perturbations, as detailed in Table~\ref{tab:pert}. All of the perturbations preserve the original meaning of the prompt, as well as the intended output. \emph{Formatting} refers to changes that modify either the structure of the prompt or the appearance of its textual content. These are rule-based perturbations which can be applied to all prompt components\footnote{Different implementations are applied to different components} and include for example: inserting extra spaces; introducing typos (e.g., ``apple'' → ``aplpe''); changing letter casing, and altering punctuation. This form of noise mimics the kind of variation found in real-world user inputs~\cite{ravichander2021noiseqachallengesetevaluation}, and has been shown to affect model performance~\cite{sclar2023quantifying}.

\emph{Paraphrasing} is an LLM-based perturbation that changes the wording of the instruction. We use the prompting method of~\citet{mizrahi2024state}, which has been shown to produce paraphrases that surface models' sensitivity. 


\emph{Context addition} perturbation adds thematically related text to the prompt without changing the gold answer or providing additional hints. While the task remains unchanged, the added content makes the prompt longer and potentially more challenging for the model \cite{levy2024same}. 

Lastly, \emph{Demonstration Editing} refers to changes to the few-shot demonstration -- namely, the number of examples, which ones are included, and their order, following~\cite{lu2021fantastically}. In addition to the general perturbation strategies, we also support task-specific features for common setups (e.g., changing enumerators in multi-answer questions). These are described in  Appendix~\ref{sec:appendix}.

\subsection{Using \name{}}






We provide a detailed overview of using \name{}. 
The package containing \name{} can be installed in the desired environment using pip:
\begin{lstlisting}[language=bash,frame=none]
    pip install promptsuite
\end{lstlisting}


\name{} transforms raw data into diverse prompt variations in four steps, as can be seen in Listing~\ref{lst:api}.



\paragraph{(1) Load datasets:} \name{} supports data from HuggingFace Datasets Library or local sources, including pandas DataFrames, JSON, and CSV files. 

\paragraph{(2) Setup template:} 
To apply the desired perturbations, \name{} requires the structure of the prompt and which dataset columns should be used in it. The \emph{Instruction}, like “Please answer the following question”, is given as a plain string. The \emph{Prompt format}, such as
\texttt{{Q: {question} A: {answer}}},
is written using Python’s f-string syntax. Each placeholder (e.g., \texttt{{question}}) must match a column name in the dataset. For example, in Listing~\ref{lst:api}, the columns are 'question' and 'answer'.

\paragraph{(3) Choose perturbations:}
Each component may be subjected to different perturbations, according to the user's choice, as described in Table~\ref{tab:pert}. These choices are added to the template setup, by specifying the name of the component and the variation. For example, in Listing~\ref{lst:api}, we choose to create LLM-based paraphrase on the instruction and apply formatting to the prompt format. To ensure maximum flexibility, users can not only modify the prompt components (i.e, instruction, prompt format, demonstrations, and instance content) but also apply alterations to any column included as a placeholder in the prompt template.

\paragraph{(4) Generate variations:} Lastly, the user can specify the number of perturbations per component. To ensure the dataset remains manageable in terms of cost and memory, users can also limit the total number of generated rows. 
Since each component supports multiple perturbations, the number of possible dataset variations grows exponentially with the number of chosen perturbations. To produce a manageable dataset size, we provide an option to randomly select a combination of the desired perturbations and apply them across the entire dataset, following the approach of~\citet{habba2025dovelargescalemultidimensionalpredictions}.

\begin{figure*}[tb!]
    \centering
    \includegraphics[width=\textwidth]{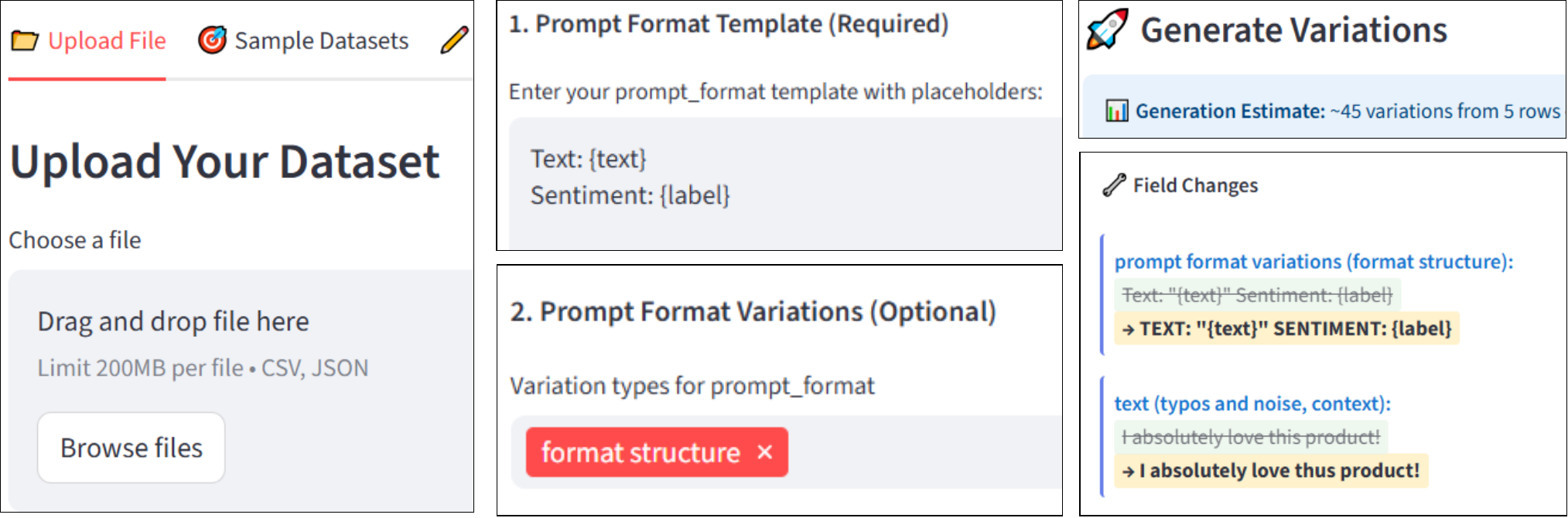}
    
\caption{\name{}'s web UI. Left-to-right: uploading a dataset; configuring the template and choosing perturbations; and generating a multi-prompt dataset. The presented example demonstrates a single prompt variation, with changes to the prompt format and instance content.}
\label{fig:web}
\end{figure*}

\paragraph{\name{} is also available via a web interface.} We offer the full capabilities of \name{} through a web UI, as illustrated in Figure~\ref{fig:web}.\footnote{\url{https://promptsuite.streamlit.app/}} Users follow the same steps described above: first, upload their single-prompt dataset, then configure the prompt components and select the desired perturbations for each component. As shown in the figure, \name{}'s web UI allows users to explore the generated variations, highlighting the changes applied to each row. The interface also provides several predefined templates for popular tasks, including multiple-choice QA, sentiment analysis, open-ended QA, and text classification, enabling a quick and easy plug-and-play setup for users who wish to automatically generate multi-prompt versions of their datasets.



\section{Evaluation}
\label{Sec:CaseStudies}

We demonstrate that PromptSuite is flexible and generalizes across a wide range of tasks by applying it to nine diverse benchmarks. Our results show that multi-prompt evaluation reveals substantial performance variance that would have been missed using a single prompt. We further assess the impact of perturbations to individual prompt components on model performance by leveraging \name{}'s modular design.

\subsection{Experimental Setup}

\paragraph{Tasks and datasets.} We evaluate PromptSuite on: (1) MMLU~\citep{hendrycks2021measuringmassivemultitasklanguage} for multiple-choice reasoning across 12 subjects; (2) GSM8K~\citep{cobbe2021training} for mathematical problem solving; (3) SST~\citep{socher-etal-2013-recursive} for sentiment analysis; (4) WMT14~\citep{bojar-etal-2014-findings} for translation across 6 language pairs (CS/HI/RU$\leftrightarrow$EN); (5) CNN/DailyMail~\citep{hermann2015teachingmachinesreadcomprehend} for summarization; (6) MuSiQue~\citep{trivedi2022musiquemultihopquestionssinglehop} for multi-hop question answering; (7) SQuAD~\citep{rajpurkar-etal-2016-squad} for reading comprehension; (8) GPQA-Diamond~\citep{rein2024gpqa} for graduate-level reasoning; and (9) HumanEval~\citep{chen2021evaluating} for code generation. 

\paragraph{Models.} We evaluate GPT-4o-mini and Llama-3.3-70B, representing closed and open-source LLMs. Temperature is set to 0 to ensure consistent and deterministic outputs. 
For code generation, we use a temperature of 0.8, since  Pass@k relies on generating multiple candidate solutions, and a non-zero temperature is essential to ensure a diverse set of outputs across multiple runs for the same prompt, as demonstrated in~\citep{chen2021evaluating}.

\paragraph{Prompt variations.} For each task, we generate variations using: paraphrasing, formatting applied to the prompt format and demonstration editing. We process 50 rows per dataset with up to 25 variations per row, resulting in approximately 1,250 evaluated prompts per task and a total of ~37,000 LLM outputs (detailed calculations in Appendix~\ref{sec:appendix}, Table~\ref{tab:combined-evaluations}, and token counts in Table~\ref{tab:token-usage}). This yields comprehensive coverage while remaining computationally tractable. 

\paragraph{Manual validation.}
To validate our results, we conducted human validation of a subset of 100 LLM-based  paraphrases. 
Two in-house annotators independently annotated all 100 samples, reaching 95\% agreement (Cohen’s k = 0.593). They judged that 96\% of the paraphrases preserved the original meaning of the instruction. The samples that were tagged as incorrect were either due to the use of a less accurate synonym (e.g., for sentiment analysis instruction, it rephrased ``sentiment'' into ``emotional tone'', which can be ambiguous, or the omission of the system message, such as ``you are an expert in QA''). 


\subsection{Results}
Below we outline interesting conclusions derived from our experiments using \name{}. 

\begin{figure*}[tb!]
    \centering
    \begin{subfigure}[b]{0.49\textwidth}
        \centering
        \includegraphics[width=\linewidth]{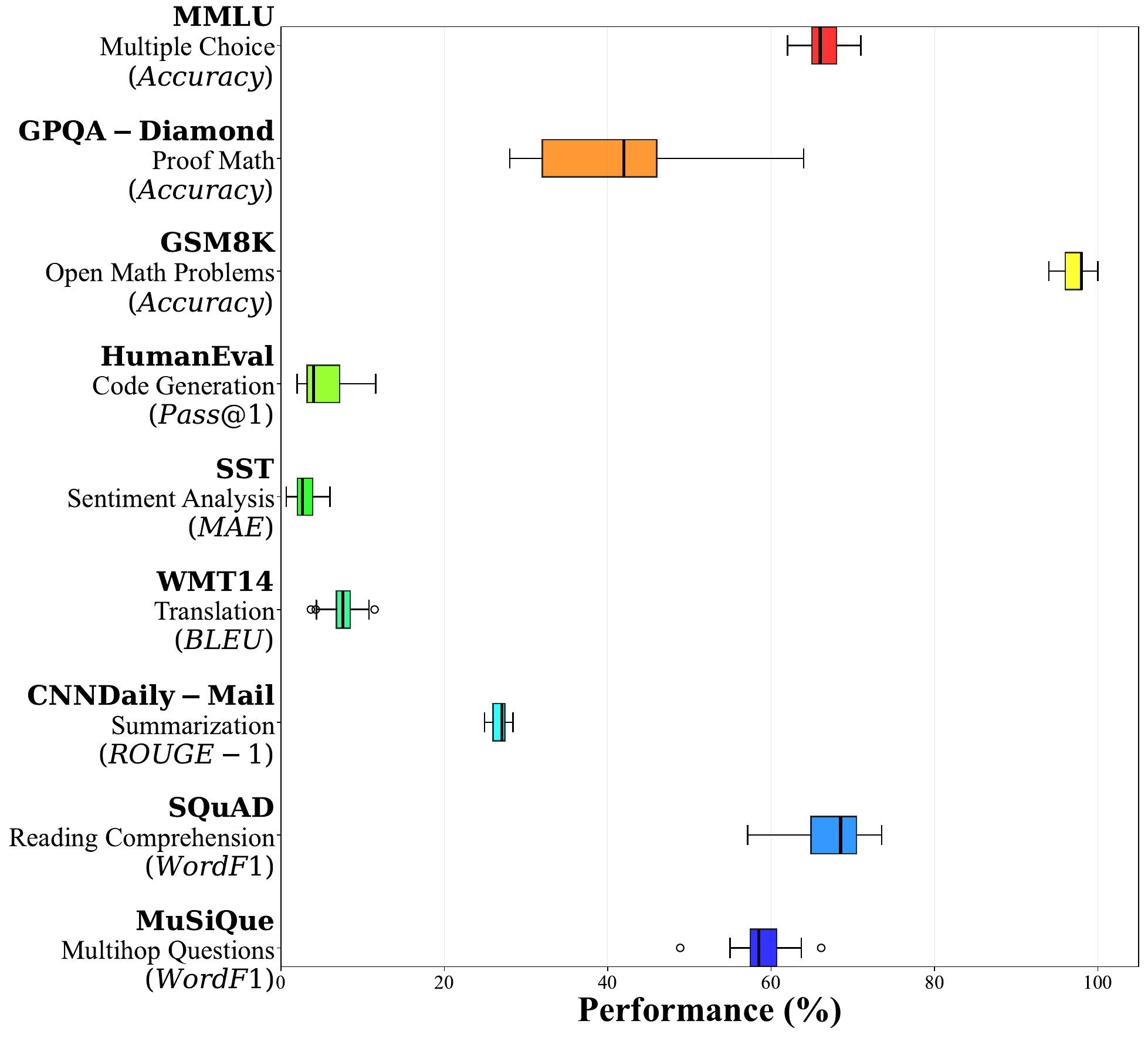}
        \caption{Llama-3.3-70B}
        \label{fig:performance_boxplot_llama}
    \end{subfigure}
    \hfill
    \begin{subfigure}[b]{0.49\textwidth}
        \centering
        \includegraphics[width=\linewidth]{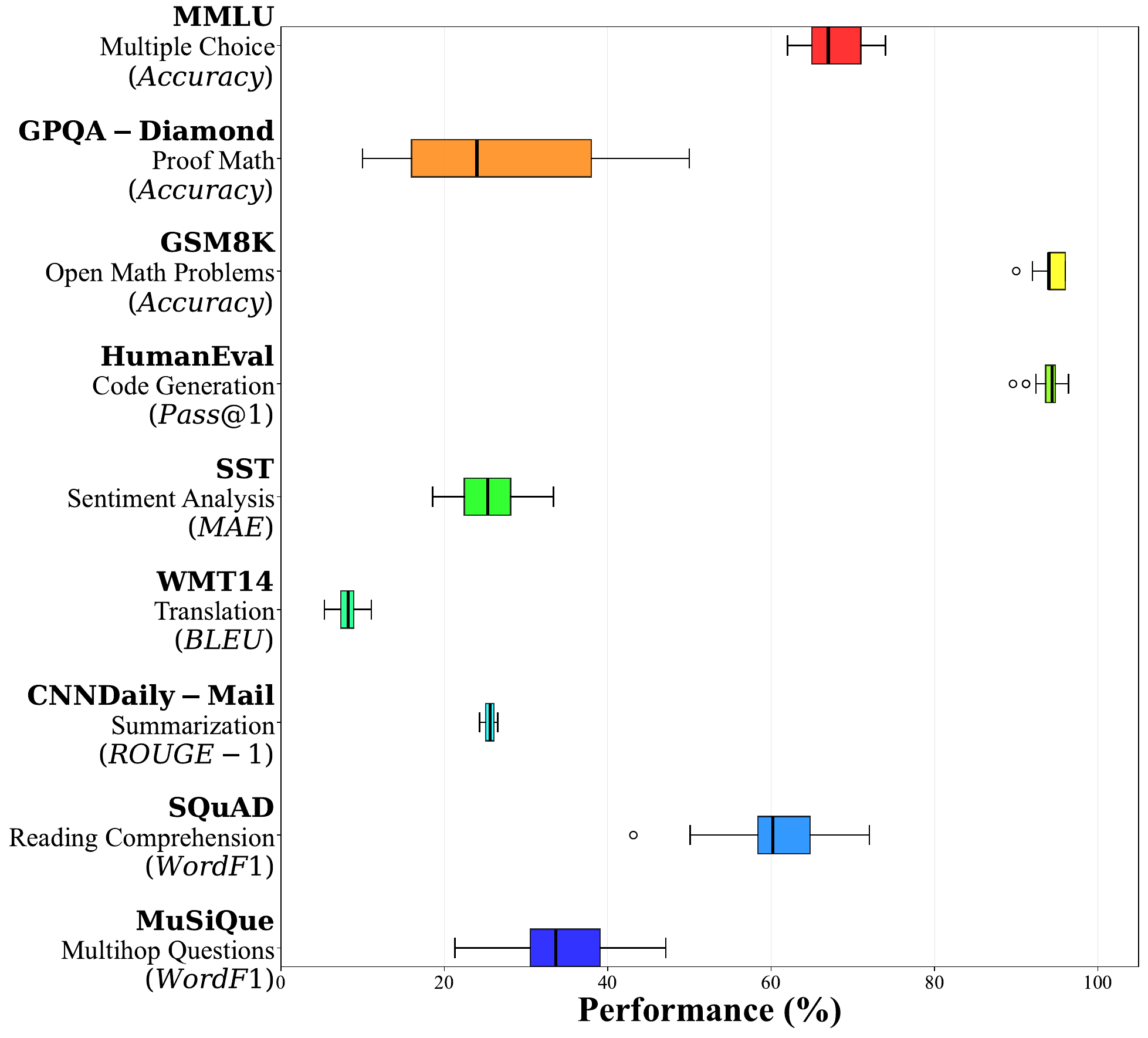}
        \caption{GPT-4o-mini}
        \label{fig:performance_boxplot_gpt}
    \end{subfigure}
    \caption{Multi-prompt evaluation results using \name{}. The boxplots illustrate variance across different prompt perturbations, revealing models' sensitivity to prompt variations and underscoring the utility of \name{} for deriving robust and meaningful evaluations of LLM capabilities.}
    \label{fig:performance_boxplot_combined}
\end{figure*}

\paragraph{Models exhibit sensitivity to the prompt perturbations across all tasks, underscoring the utility of \name{}.}
Figures~\ref{fig:performance_boxplot_llama} and~\ref{fig:performance_boxplot_gpt} show performance distributions across prompt variations for \llama{} and \gpt{}, respectively. We observe substantial variability. For instance, on GPQA-Diamond, \gpt{}'s accuracy ranges from 20\% to 50\% across variations. This variance is particularly striking when compared to typical performance differences between competing models, which often amount to just a few percentage points \cite{lior2025reliableeval}. The consistency of this pattern across diverse tasks demonstrates that prompt sensitivity is not limited to specific domains but represents a general challenge in LLM evaluation.




\begin{figure}[t]
    \centering
    \includegraphics[width=\linewidth]{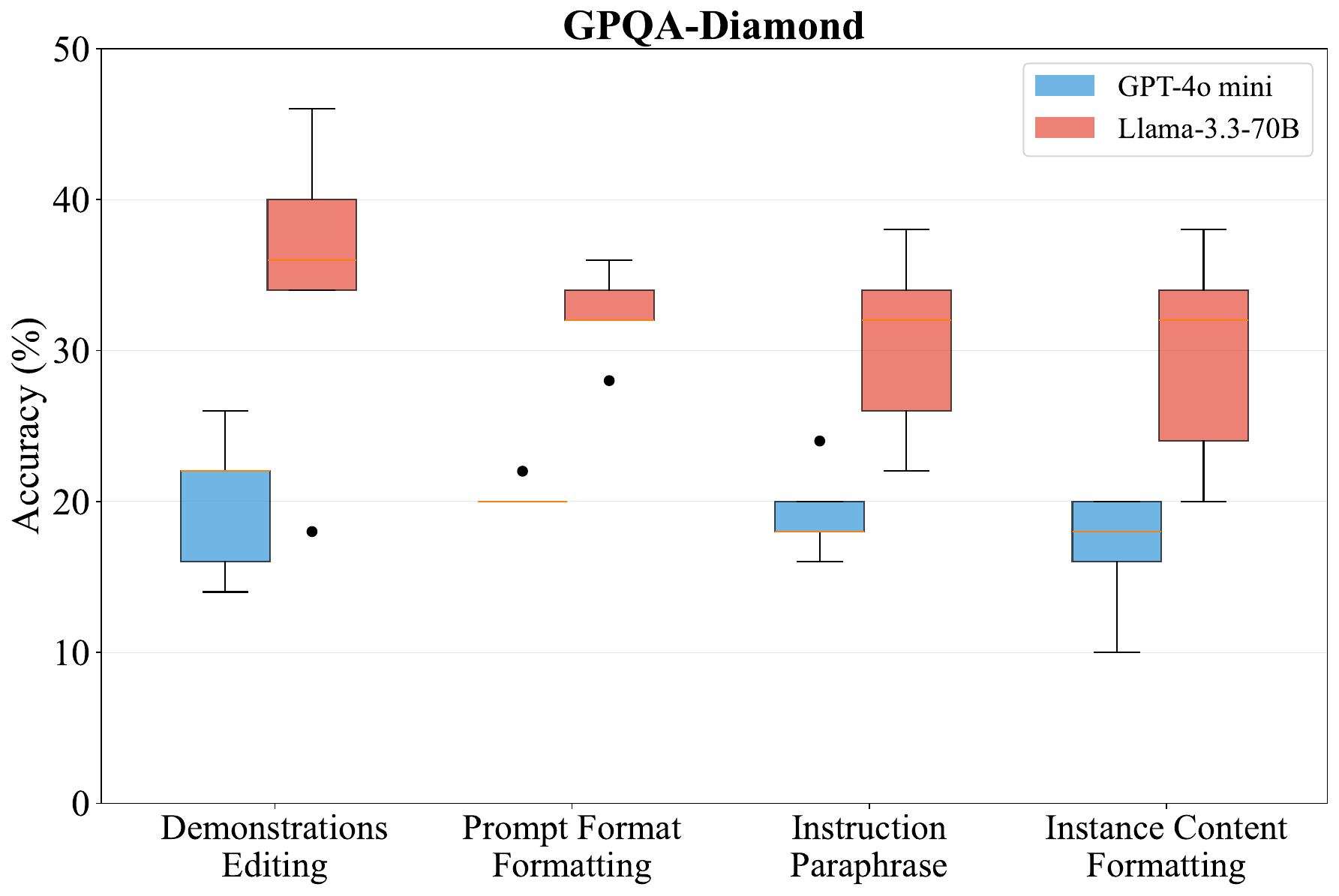}
    \caption{
    Analysis of how perturbations to individual prompt components affect model sensitivity on GPQA-Diamond. Each boxplot represents an experiment in which a single prompt component was varied while all others remained fixed.
    }
    \label{fig:component-analysis-gpqa}
\end{figure}

\paragraph{\name{}'s modularity enables systematic ablation, showing that the impact of prompt component perturbations varies across tasks and models.}
\name{}'s modularity allows a systematic ablation study that assesses the impact of changes in specific prompt components on model performance. We perturb one prompt component at a time to measure its specific effect, testing instruction paraphrasing, formatting applied to either the prompt format or the instance content and demonstrations editing. 
 We conduct this experiment on GPQA-Diamond, SQuAD and GSM8K. Each perturbation  was evaluated on 50 examples with 20 variations per example, yielding 1000 evaluated prompts per component-task combination for each model.
 Figure~\ref{fig:component-analysis-gpqa}
 presents the performance distributions across perturbation types in GPQA-Diamond. For example, we observe that demonstration editing caused high variance in \llama{}'s performance, whereas for other tasks (Figure~\ref{fig:componentsquadgs} in the Appendix), demonstration editing had almost no effect on \llama{}'s performance. Similarly, for \gpt{}, prompt formatting had almost no effect on GPQA-Diamond (Figure~\ref{fig:component-analysis-gpqa}), but showed a more significant effect for SQuAD (Figure~\ref{fig:componentsquadgs}). 
 These inconsistencies across models and tasks underscore the importance of a flexible and modular framework like \name{}, which enables systematic analysis of prompt component effects.
For example, practitioners
  can leverage \name{} for a more efficient evaluation strategy, by first experimenting on a small subset of the data to identify the most influential prompt components, and then conducting a focused large-scale evaluation that concentrates only on the perturbations with the most significant impact.

\section{Related Work}
A few existing frameworks support a subset of \name{}'s capabilities. To the best of our knowledge, they are task-specific and require manual control over input variations. NL-Augmenter~\cite{dhole2021nl} is a crowd-sourced repository for perturbations. While it provides a wide range of task-specific transformations and filters, it operates solely on the input data and does not account for instructions or templates, both of which are critical for robust few-shot evaluation. Unitxt~\cite{bandel2024unitxt} is a library for data preparation and evaluation, with some tasks supporting a limited number of data alterations. Prompt-Agnostic Fine-Tuning (PAFT)~\cite{wei2025paftpromptagnosticfinetuning} also seeks to reduce prompt sensitivity by generating diverse prompts, but incorporates them during finetuning rather than at evaluation time.
\section{Conclusion}
We introduce \name{}, a framework that generates prompt variations needed for multi-prompt evaluation. It is flexible, uses a modular design for controlled perturbations, and is easily extensible. 
Through case studies, we show that the variations generated by \name{} are sufficient to test model sensitivity.

\section{Limitations}
While \name{} provides a general, task-agnostic framework for multi-prompt evaluation, this generality comes with a tradeoff. Some tasks may involve specific variations or prompt structures that \name{} does not currently support. Specifically, this limitation arises in cases where evaluation is not straightforward (e.g., multi-turn chat).
To mitigate this limitation, we designed our code to be easily extensible, allowing users to add additional prompt components or variations as needed.


\section*{Acknowledgments}
This research was also supported by the Ministry of Innovation, Science \& Technology, Israel (Grant No. 0008239). We would like to thank our colleagues Asaf Yehudai, Hillel Darshan, Daniel Nisnevich, Nitzan Barzilay, and Michael Hassid for their contributions to our initial prototype development, which served as an inspiration for this work. We would also like to thank Omer Kidron for his assistance in recording the demonstration video.

\bibliography{custom}

\appendix

\section{Example Appendix}

\subsection{Task-Specific Perturbations}
For multiple choice questions, multi-document or any tasks that includes a list as input we offer perturbations of said list, presented in Table~\ref{tab:pert2}. This includes support for enumerating the list items, as well as shuffling the order (while changing the gold answer accordingly, if applicable).

\begin{table*}[tb!]
\resizebox{\textwidth}{!}{%
\small
\centering
\begin{tabular}{@{}p{3cm}p{2cm}p{7cm}p{2cm}@{}}

\toprule
\textbf{Perturbation Type} & \textbf{Applicable Fields}  & \textbf{Description} \\
\midrule
Enumerate & lists, comma separated values & adds enumeration to a specified field, such as multiple-choice options, by prepending each item with a number or letter (e.g., 1., A., a.). Following~\cite{habba2025dovelargescalemultidimensionalpredictions}
\\
 \hline
Shuffle	 & lists & shuffles the items in a list and updates the gold field to reflect the new index of the correct answer. For example, if the correct answer was originally at position B and is moved to position C after shuffling, the gold label is updated accordingly. Following~\cite{habba2025dovelargescalemultidimensionalpredictions}
 \\ 

\bottomrule

\end{tabular}
}

\caption{Task-specific  perturbation types in \name{}. The "Applicable Fields" column indicates which types of data column the perturbation works on.}
\label{tab:pert2}
\end{table*}

\begin{table*}[ht]
\centering
\small
\begin{tabular}{@{}lrrr@{}}
\toprule
\textbf{Benchmark/Task} & \textbf{Questions} & \textbf{Variations} & \textbf{Total (Q × V)} \\
\midrule
MMLU Multiple Choice \newline (12 subjects, 10 questions each) & 120 & 50 & 6000 \\
GSM8K Open Math Problems & 50 & 25 & 1250 \\
HumanEval Code Generation & 50 & 25 & 1250 \\
SST Sentiment Analysis & 50 & 50 & 2500 \\
WMT14 Translation \newline (CS/HI/RU $\leftrightarrow$ EN) & 60 & 50 & 3000 \\
CNN-DailyMail Summarization & 50 & 25 & 1250 \\
MuSiQue Multihop Questions & 50 & 25 & 1250 \\
SQuAD Reading Comprehension & 50 & 25 & 1250 \\
GPQA--Diamond Google-Proof Math & 50 & 25 & 1250 \\
\midrule
\textbf{Total across both models} & -- & -- & \textbf{37,375} \\
\bottomrule
\end{tabular}
\caption{
Number of evaluated examples per benchmark. 
Each row indicates the number of base questions and variations, with the total computed as their product.
\textbf{Note:} Values shown reflect the GPT-4o mini configuration. For LLaMA-3-3.7B, the MuSiQue dataset included only 25 base questions (instead of 50) due to limited context window constraints, yielding a total of 625 examples for that task. Total reflects the combined number of evaluations across both models.
}
\label{tab:combined-evaluations}
\end{table*}

\begin{figure*}[tb!]
    \centering
    \begin{subfigure}[b]{0.49\textwidth}
        \centering
        \includegraphics[width=\linewidth]{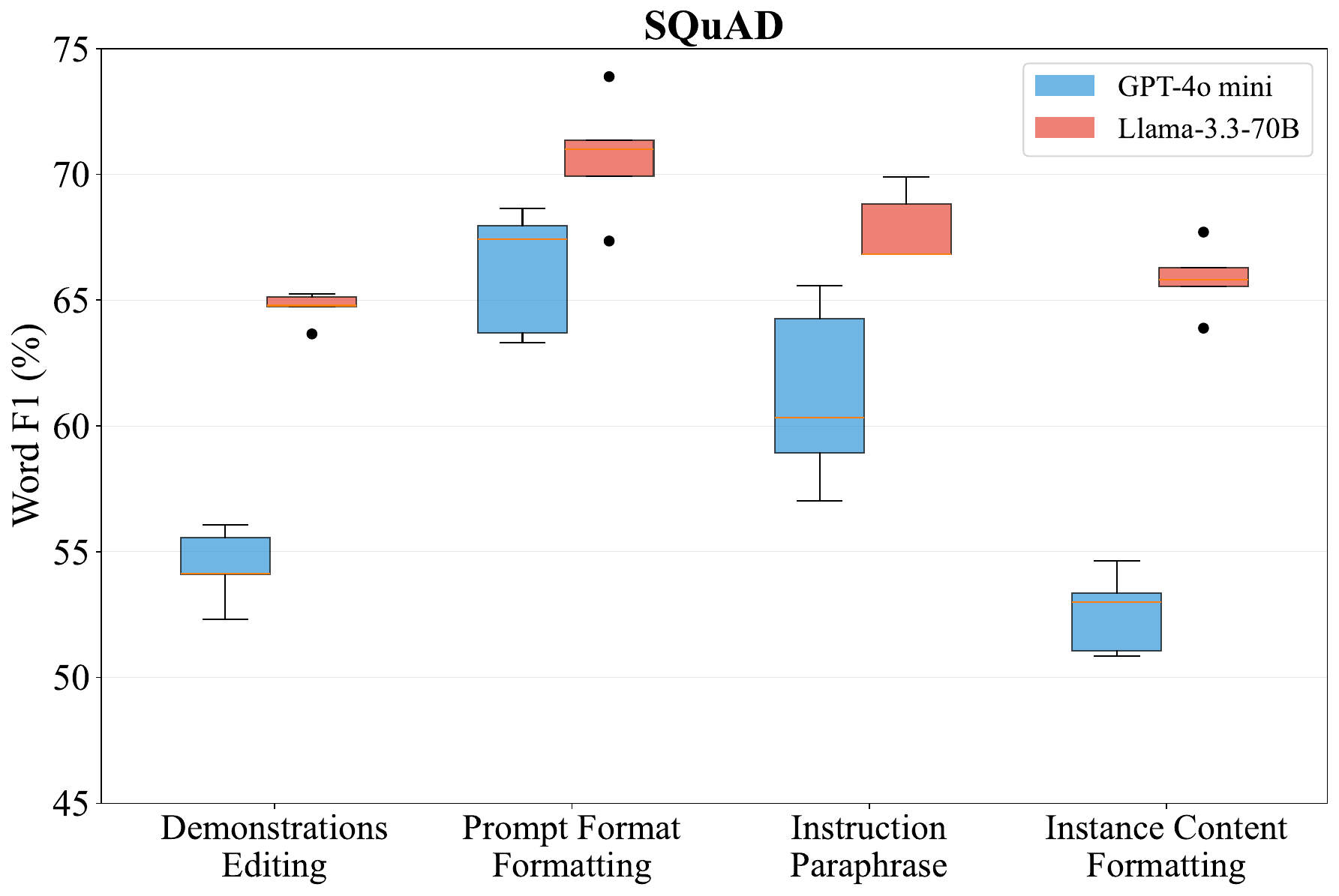}
        \label{fig:component-analysis-squad}
    \end{subfigure}
    \hfill
    \begin{subfigure}[b]{0.49\textwidth}
        \centering
        \includegraphics[width=\linewidth]{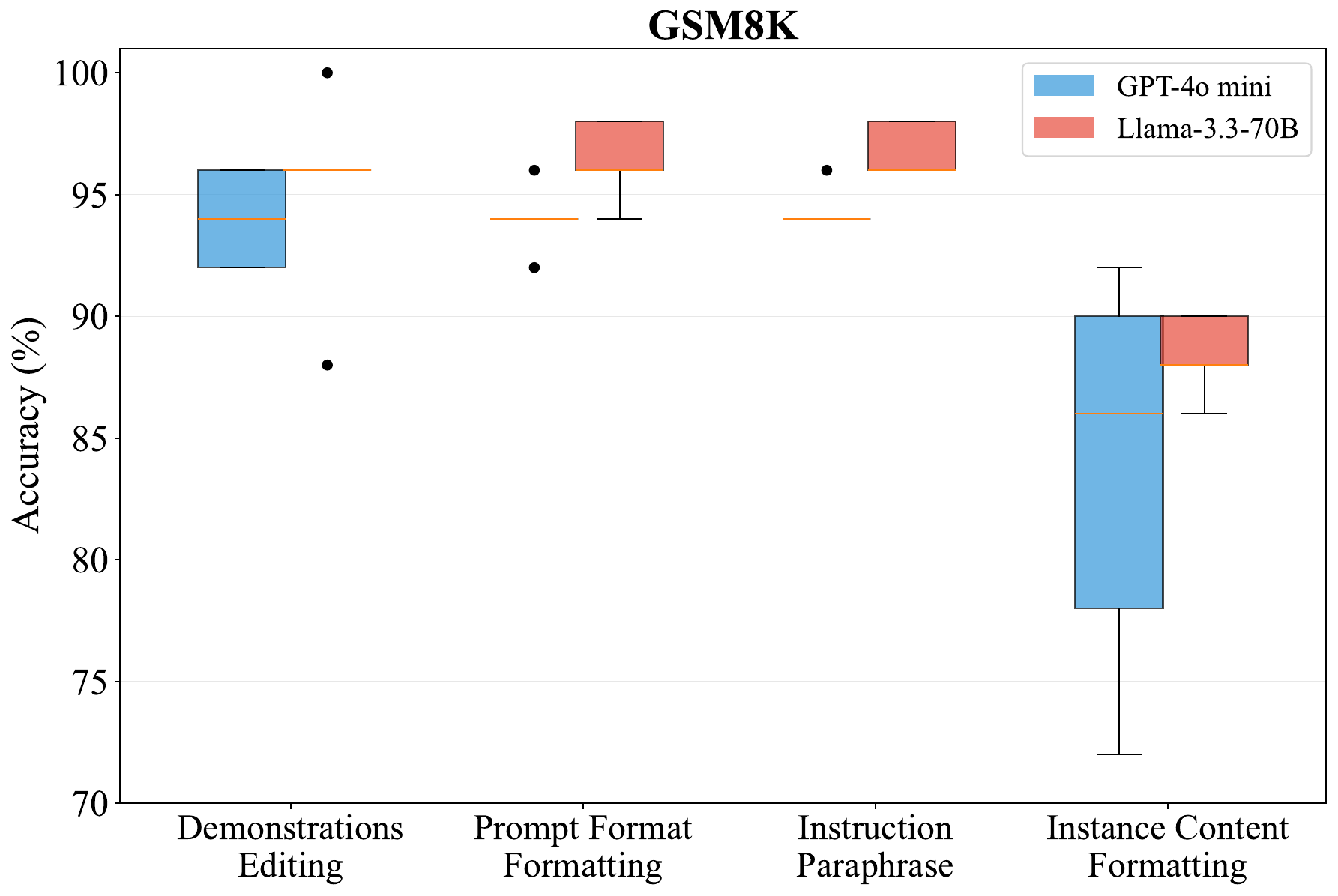}
        \label{fig:component-analysis-gsm8k}
    \end{subfigure}
    \caption{Analysis of how perturbations to individual prompt components affect model sensitivity on SQuAD and GSM8K. Each boxplot represents an experiment in which a single prompt component was varied while all others remained fixed.}
    \label{fig:componentsquadgs}
\end{figure*}



\begin{table*}[ht]
\centering
\small
\begin{tabular}{@{}lrrrr@{}}
\toprule
\textbf{Benchmark/Task} 
& \multicolumn{2}{c}{\textbf{Input Tokens}} 
& \multicolumn{2}{c}{\textbf{Output Tokens}} \\

 & GPT-4o mini & Llama-3.3 70B & GPT-4o mini & Llama-3.3 70B \\
\midrule
MMLU Multiple Choice & 1389791 & 1391104 & 68403  & 173011 \\
GSM8K Open Math Problems & 722880  & 731987  & 223225 & 135769 \\
HumanEval Code Generation & 7259990 & 7228940 & 285569 & 285313 \\
SST Sentiment Analysis & 479750  & 487721  & 10934  & 108987 \\
WMT14 Translation & 409989  & 422709  & 34478  & 95237 \\
CNN-DailyMail Summarization & 3888319 & 3940895 & 167329 & 155074 \\
MuSiQue Multihop Questions & 8586023 & 8102151 & 17371  & 14525 \\
SQuAD Reading Comprehension & 1279578 & 1288793 & 10099  & 9787 \\
GPQA--Diamond Google-Proof Math & 1658817 & 1679053 & 384586 & 461600 \\
\bottomrule
\end{tabular}
\caption{
Token usage per benchmark across GPT-4o mini and Llama-3.3 70B. 
The table shows the number of input and output tokens consumed for each benchmark.
}
\label{tab:token-usage}
\end{table*}
\label{sec:appendix}

\end{document}